\documentclass{Interspeech}
\usepackage{subcaption}
\usepackage{hyperref}

\usepackage{subcaption}
\usepackage{fancybox}
\usepackage{tikz}
\usepackage{listings}

\definecolor{codegray}{gray}{0.9}

\lstdefinestyle{pythonstyle}{
    backgroundcolor=\color{codegray},   
    language=Python,
    basicstyle=\ttfamily\footnotesize,
    keywordstyle=\color{blue},
    stringstyle=\color{red},    
    breaklines=true,
    frame=single,
    keepspaces=true,
    showstringspaces=false,
}

\lstdefinestyle{mystyle}{
  language=Python,
  basicstyle=\ttfamily\footnotesize,
  keywordstyle=\color{blue},
  commentstyle=\color{gray},
  stringstyle=\color{gray},
  backgroundcolor=\color{white},
  numberstyle=\tiny\color{gray},
  numbersep=1pt,
  frame=single,
  breaklines=true,
  breakindent=0pt,
  showstringspaces=false,
  captionpos=b,
  xleftmargin=0pt,
  framexleftmargin=0pt,
  framexrightmargin=0pt  
}

\interspeechcameraready

\title{From Words to Waves: Analyzing Concept Formation in Speech and Text-Based Foundation Models}

\author{As{\i}m}{Ersoy}
\author{Basel}{Mousi}
\author{Shammur}{Chowdhury}
\author{Firoj}{Alam}
\author{Fahim}{Dalvi}
\author{Nadir}{Durrani}

\affiliation[nocounter]{Qatar Computing Research Institute, HBKU}{Qatar}

\email{shchowdhury@hbku.edu.qa, ndurrani@hbku.edu.qa}


\usepackage{comment}

\begin{document}

\maketitle

\begin{abstract}

The emergence of large language models has demonstrated that systems trained solely on text can acquire extensive world knowledge, develop reasoning capabilities, and internalize abstract semantic concepts--showcasing properties that can be associated with general intelligence. This raises an intriguing question: \textit{Do such concepts emerge in models trained on other modalities, such as speech?} Furthermore, when models are trained jointly on multiple modalities: \textit{Do they develop a richer, more structured semantic understanding?} 
To explore this, we analyze the conceptual structures learned by speech and textual models both individually and jointly. We employ Latent Concept Analysis, an unsupervised method for uncovering and interpreting latent representations in neural networks, to examine how semantic abstractions form across modalities. To support reproducibility, we have released our \textbf{code}\footnote{\href{https://github.com/shammur/MultimodalXplain}{https://github.com/shammur/MultimodalXplain}} along with a curated \textbf{audio version of the SST-2 dataset}\footnote{\href{https://huggingface.co/collections/QCRI/multimodalxplain-6839bbe6fc98a0b221dc42bb}{https://huggingface.co/collections/QCRI/multimodalxplain-6839bbe6fc98a0b221dc42bb}} for public access.  
\end{abstract}

\section{Introduction}


Recent advances in artificial intelligence have led to the development of large neural models capable of processing and generating language, vision, and speech \cite{bubeck2023sparks, liang2022holistic, touvron2023llama, chu2024qwen2, zhao2025humanomni}. Among these, large language models (LLMs) have demonstrated emergent capabilities once thought to require human intelligence. From commonsense reasoning to medical diagnosis and legal analysis, these models have continuously pushed the boundaries of AI-driven understanding and decision-making \cite{bubeck2023sparks,jeblick2022chatgpt}. Their ability to internalize abstract concepts, perform multi-step reasoning, and apply knowledge in novel contexts has fueled discussions on their potential trajectory toward artificial general intelligence (AGI).

However, a fundamental question remains:  \textit{Are these emergent capabilities unique to text-based models, or do similar properties arise in models trained on other modalities, such as speech?} Furthermore, when models are trained on multiple modalities, \textit{do they converge toward a shared semantic space that facilitates conceptual abstraction}, as proposed by the \emph{Semantic Hub Hypothesis} \cite{Patterson2007WhereDY}?
To explore these questions, we analyze the conceptual structures learned by speech models and compare them to text-based models and jointly trained multimodal systems. We employ Latent Concept Analysis (LCA) \cite{dalvi2022discovering}, to uncover and compare the abstract concepts formed within these models. We align the discovered concepts with predefined taxonomies, 
enabling a structured comparison of semantic representations across modalities.

In our study, we investigate unimodal models (HuBERT for speech and BERT for text) and multimodal models (Seamless M4T and SpeechT5) to assess their ability to capture linguistic knowledge. We evaluate these models using core linguistic tasks like part-of-speech tagging, chunking, and semantic analysis, as well as sentiment analysis, to compare their ability to align with human-defined concepts and task-specific representations. 
Our study addresses the following research questions: 
\begin{itemize}
    \item \textbf{Question:} How do conceptual structures in speech models differ from those in text and multimodal models?  \newline
          \textbf{Finding:} The alignment patterns reveal that text models directly encode linguistic taxonomies from early layers, while speech models gradually transition from acoustic to linguistic representations. Multimodal models like SpeechT5 show unique alignment due to cross-modal training.
          
    \item \textbf{Question:} To what extent do different modalities yield shared or modality-specific semantic representations?  
    \newline
          \textbf{Finding:} Speech models allocate less capacity to linguistic and semantic taxonomies, focusing more on speech-specific features like phonetics. In contrast, text models, which operate on tokenized inputs, develop more structured and deep linguistic representations. This highlights the disparity in how each modality internalizes semantic structures.
\end{itemize}



\section{Methodology}

Our methodology is designed to uncover and compare the latent conceptual structures emerging within speech, text, and multimodal foundation
models. To achieve this, we employ Latent Concept Analysis (LCA) \cite{dalvi2022discovering}, an unsupervised approach that enables the discovery and interpretation of abstract representations learned by neural networks. Our approach consists of two stages: \textbf{concept discovery} and \textbf{concept alignment}. 

\subsection{Concept Discovery}
\label{subsec:concept_discovery}

\begin{figure*}
     \centering
     \begin{subfigure}[b]{0.24\textwidth}
         \centering
         \includegraphics[width=\textwidth]{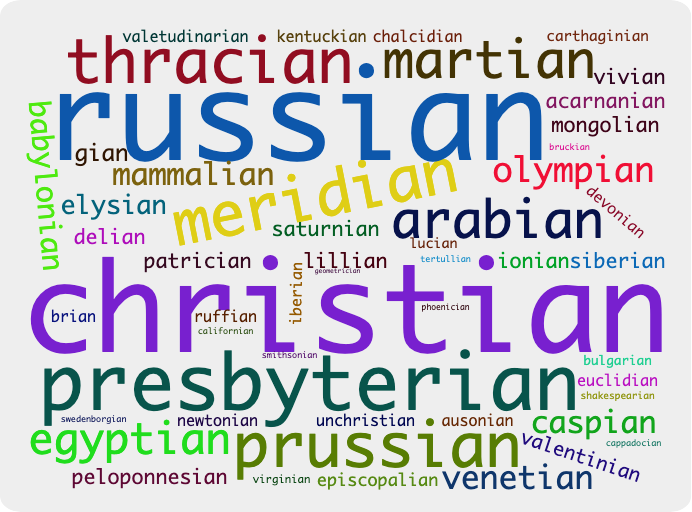}
         \caption{Nationalities and Ethnicities}
         \label{fig:bertL6C66}
     \end{subfigure}
     \begin{subfigure}[b]{0.24\textwidth}
         \centering
         \includegraphics[width=\textwidth]{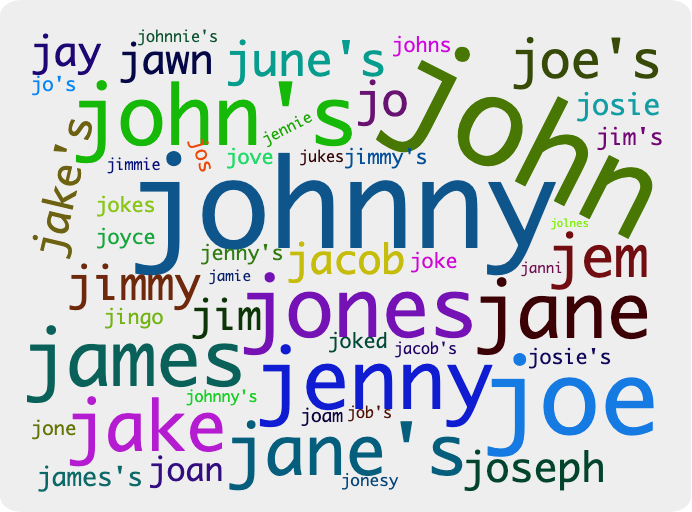}
         \caption{Names Starting with \textipa{/dZ/}}
         \label{fig:HubertL12c141}
     \end{subfigure}
     \begin{subfigure}[b]{0.24\textwidth}
         \centering
         \includegraphics[width=\textwidth]{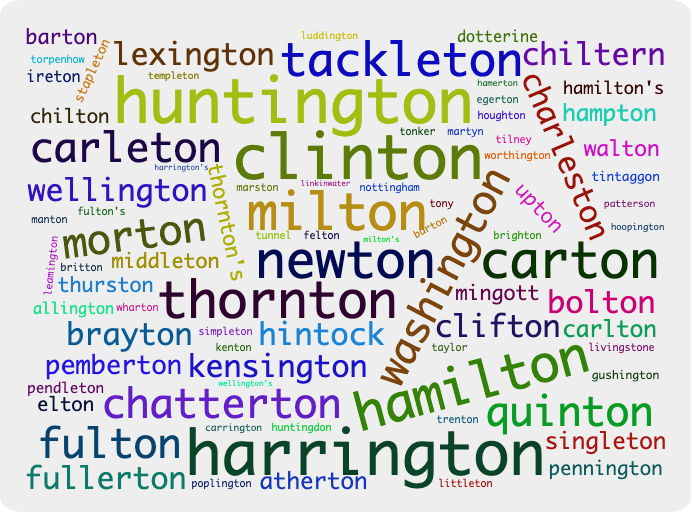}
         \caption{NEs ending in the \textipa{/t@n/} sound}
         \label{fig:hubertL12C216}
     \end{subfigure}
     \begin{subfigure}[b]{0.24\textwidth}
         \centering
         \includegraphics[width=\textwidth]{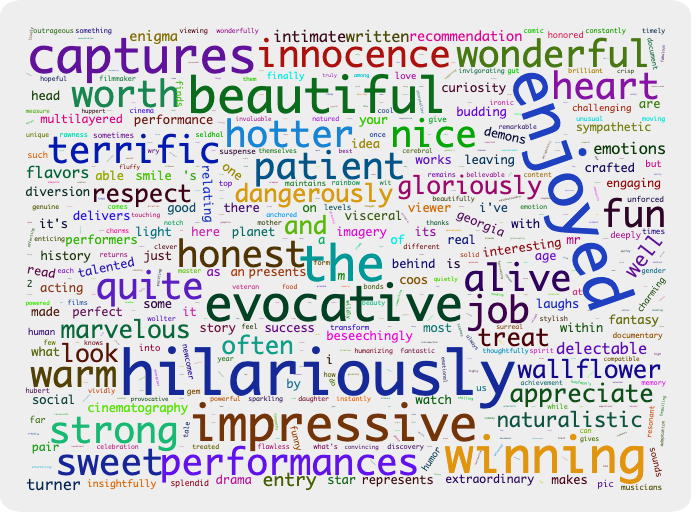}
         \caption{Positive Polarity Concept}
         \label{fig:BertL22C0}
     \end{subfigure}     
    \caption{Sample latent concepts from different models. Figures \ref{fig:bertL6C66} \& \ref{fig:BertL22C0} show BERT concepts; \ref{fig:HubertL12c141} \& \ref{fig:hubertL12C216} show HuBERT concepts.}
    \label{fig:SampleConceptsLearned}
\end{figure*}

Contextualized representations learned in 
foundation models capture latent conceptual structures that can be interpreted through clustering methods. Our investigation expands upon the work done in discovering latent ontologies in contextualized representations \cite{michael-etal-2020-asking}. We extract contextualized representations from unimodal models 
and multimodal models, each with \(L\) layers (\(l_1, l_2, \ldots, l_L\)), and cluster them to obtain encoded concepts. In this context, a concept refers to a set of linguistic units such as words, phonemes, or acoustic patterns, grouped based on lexical, semantic, syntactic, morphological, or phonetic relationships. Figure \ref{fig:SampleConceptsLearned} showcases concepts within the latent space of the different models, wherein word representations are arranged based on distinct linguistic and task-specific concepts.

\noindent\textbf{Textual Input.}  
Given a textual utterance \(\mathcal{U} = [w_1, \dots, w_N]\), we extract contextual embeddings at layer \(l\): $\mathcal{U} \xrightarrow{\mathbf{M}_t^l} \mathbf{\Phi}^l = [\mathbf{\phi}^l_1, \dots, \mathbf{\phi}^l_N]$
where \(\mathbf{\phi}_i^l\) is the embedding of \(w_i\) at layer \(l\).

\noindent\textbf{Speech Input.}  
For the corresponding speech utterance \(\mathcal{X} = [x_1, x_2, \ldots, x_T]\), consisting of \(T\) frames, we extract frame-level representations \(\mathbf{\Psi}^l\) at each layer. Since our analysis focuses on word-based concepts, we derive word-level representations\footnote{Also known as the acoustic word embeddings \cite{sanabria2023acoustic}} \(\mathbf{\Psi}_w\) by averaging frame embeddings within the word boundary \([t_{\text{start}}, t_{\text{end}}]\) \cite{sanabria2023acoustic}.  
Word boundaries are obtained using the Montreal Forced Aligner.\footnote{\href{https://github.com/MontrealCorpusTools/Montreal-Forced-Aligner}{https://github.com/MontrealCorpusTools/Montreal-Forced-Aligner}}

\subsection{Concept Alignment}
\label{subsec:concept_annotation}



Encoded concepts capture latent relationships among words within a cluster, encompassing phonetic, lexical, syntactic, semantic, and task- or modality-specific patterns. To systematically interpret these concepts, we employ an alignment metric proposed by \cite{hawasly-etal-2024-scaling}, which maps the discovered concepts to structured linguistic ontologies using predefined taxonomies.

Let \( \mathcal{C_L} = \{ C_{l_{1}}, C_{l_{2}}, \dots, C_{l_{n}} \} \) be the set of linguistic concepts (e.g., parts-of-speech tags of words), and \( \mathcal{C_E} = \{ C_{e_{1}}, C_{e_{2}}, \dots, C_{e_{m}} \} \) be the set of encoded concepts discovered within neural language models. We define their $\theta$-alignment as follows:  
\vspace{-2mm}
\begin{equation}  
\lambda_{\theta}(\mathcal{E}, \mathcal{L}) = \frac{1}{2} \left( \frac{\sum_{\mathcal{E}}\alpha_{\theta}(C_e)}{|\mathcal{C_E}|} + \frac{\sum_{\mathcal{H}}\kappa_{\theta}(C_l)}{|\mathcal{C_L}|} \right) \times 100
\nonumber
\end{equation}  
\vspace{-2mm}
where alignment $\alpha_{\theta}(C_e)$ and coverage $\kappa_{\theta}(C_l)$ are defined as
\begin{equation}  
\alpha_{\theta}(C_e) =  
\begin{cases}  
    1, & \text{if } \exists C_l \in \mathcal{C_L} \text{ such that } \frac{|C_e \cap C_l| }{|C_e|} \geq \theta  \\  
    0, & \text{otherwise}
    \nonumber
\end{cases}  
\end{equation}  
\begin{equation}  
\kappa_{\theta}(C_l) =  
\begin{cases}  
    1, & \text{if } \exists C_e \in \mathcal{C_E} \text{ such that } \frac{|C_e \cap C_l| }{|C_e|} \geq \theta  \\  
    0, & \text{otherwise}  
\end{cases}  
\nonumber
\end{equation}  

The alignment term measures how many discovered concepts align with the categories of the underlying taxonomy, while the coverage term assesses how many linguistic concepts from a given taxonomy appear in the discovered clusters.

\begin{figure*}
     \centering
     \begin{subfigure}[b]{0.5\textwidth}
         \centering
         \includegraphics[width=\textwidth]{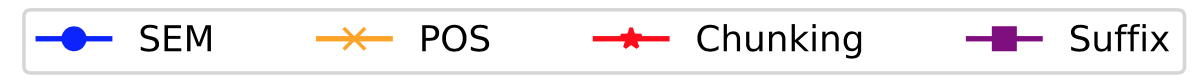}
     \end{subfigure}
     
     \begin{subfigure}[b]{0.32\textwidth}
         \centering
         \includegraphics[width=\textwidth]{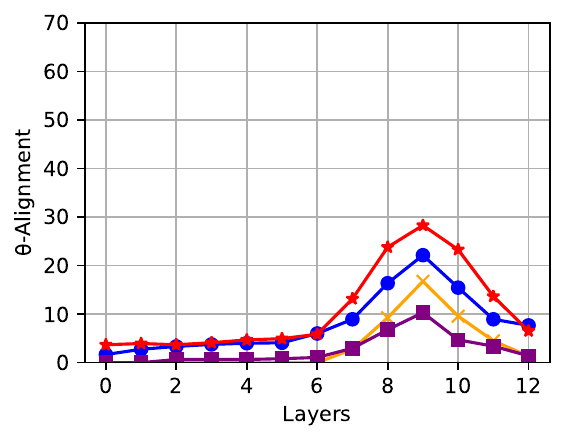}
         \caption{HuBERT}
         \label{fig:hubert-alignment}
     \end{subfigure}
     \begin{subfigure}[b]{0.32\textwidth}
         \centering
         \includegraphics[width=\textwidth]{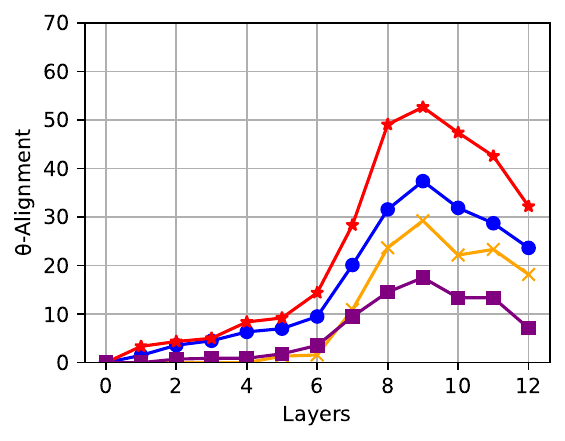}
         \caption{Seamless -- Speech}
         \label{fig:seamless-speech-encoder-alignment}
     \end{subfigure}
      \begin{subfigure}[b]{0.32\textwidth}
          \centering
          \includegraphics[width=\textwidth]{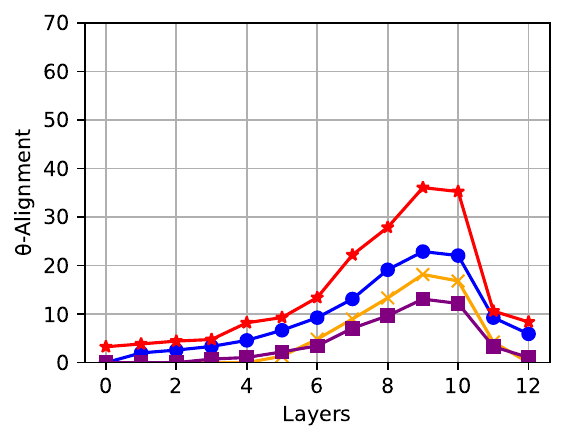}
          \caption{SpeechT5 -- Speech}
          \label{fig:Speech-T5-Speech}
     \end{subfigure}
      \begin{subfigure}[b]{0.32\textwidth}
         \centering
         \includegraphics[width=\textwidth]{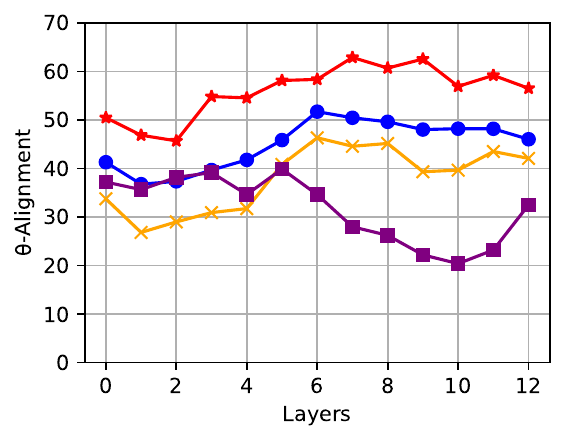}
         \caption{BERT}
         \label{fig:bert-alignment}
     \end{subfigure}
    \begin{subfigure}[b]{0.32\textwidth}
         \centering
         \includegraphics[width=\textwidth]{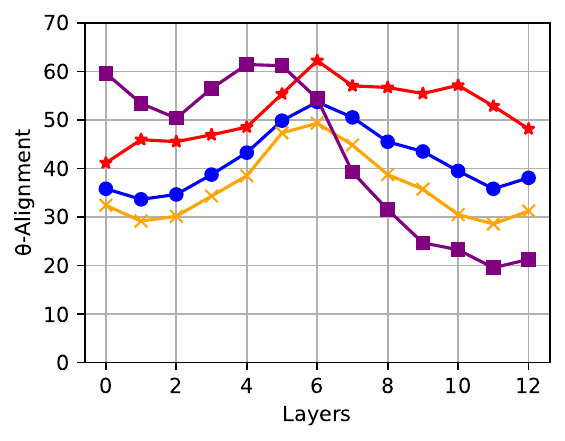}
         \caption{Seamless -- Text}
         \label{fig:seamless-text-encoder-alignment}
     \end{subfigure}
      \begin{subfigure}[b]{0.32\textwidth}
         \centering
         \includegraphics[width=\textwidth]{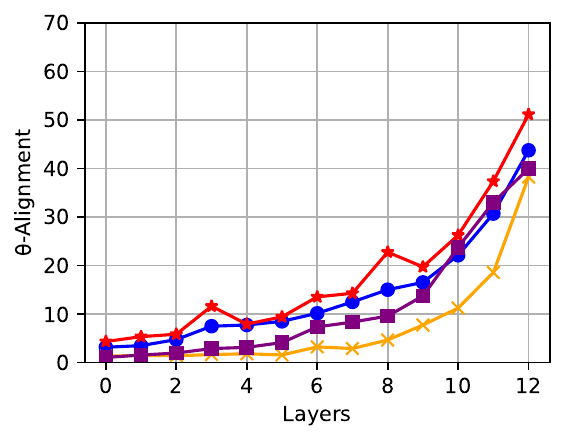}
         \caption{SpeechT5 -- Text}
         \label{fig:SpeechT5-text}
     \end{subfigure}
     \vspace{-0.3cm}
     \caption{Layer-wise concept alignment. Y-axis shows \% of aligned concepts (using $\theta{=}0.9$) per encoder layer (X-axis).
     }
     \vspace{-0.3cm}
    \label{fig:models-alignment}
\end{figure*}

\section{Experimental Setup}
\label{sec:setup}



\textbf{Models.} We investigate both unimodal and multimodal models, focusing on HuBERT, BERT, Seamless M4T, and SpeechT5. HuBERT \cite{hsu2021hubert} is a self-supervised speech model that excels at learning speech representations through masked prediction of quantized acoustic features. BERT \cite{devlin-etal-2019-bert} is a widely used pre-trained text model that captures rich contextual information in language through bidirectional transformers. 
We explore two multimodal models: Seamless M4T \cite{seamlessm4t}, which combines speech and text in a shared decoder, and SpeechT5 \cite{speecht5}, an extension of T5 that jointly learns speech and text for tasks like speech-to-text and text-to-speech. These models represent a spectrum of approaches to analyze representations learned across different modalities.









\noindent \textbf{Tasks.} We conducted experiments using traditional taxonomies designed to capture core linguistic concepts. These include word morphology, represented by part-of-speech tagging  \cite{marcus-etal-1993-building}; syntax, explored through chunking tagging \cite{tjong-kim-sang-buchholz-2000-introduction}; 
and semantics, examined through Parallel Meaning Bank annotations \cite{abzianidze-EtAl:2017:EACLshort}. We trained sequence taggers for each of these tasks and annotated the corresponding training data. Each core linguistic task represents a human-defined concept, which we align with encoded representations to assess how linguistic knowledge is structured in the model's latent space. We also used the shallow lexical concept such as suffixation. 

We also compared encoded concepts with task-specific concepts in sentiment analysis \cite{socher-etal-2013-recursive} using the alignment function to measure affinity. For the \textit{positive sentiment concept}, we define \( C_{sst}(+ve) \) as the set of words appearing exclusively in positively labeled sentences. An encoded concept \( C_{e} \) is considered aligned with \( C_{sst}(+ve) \) if a threshold \( \theta \) of its words also appear in positive sentences, with each word represented by its contextualized embedding. The same process applies to \( C_{sst}(-ve) \) to identify negative polarity concepts.




\noindent \textbf{Data.} We used two datasets for concept analysis: LibriSpeech \cite{panayotov2015librispeech}, a large-scale read-speech corpus with diverse speakers, and Stanford Sentiment Treebankv2 (SST2) \cite{socher-etal-2013-recursive}, a text-only sentiment classification dataset. \textbf{We extended SST2 into the speech domain (\textbf{SST2-audio}) using XTTSv2},\footnote{\href{https://docs.coqui.ai/en/latest/models/xtts.html}{https://docs.coqui.ai/en/latest/models/xtts.html}} a controllable TTS model, to generate high-quality audio. SST2-audio preserves sentiment polarity while minimizing speaker and environmental variability, enabling analysis of semantic concepts in speech models. 


\noindent \textbf{Concept Discovery and Annotation.} We perform a forward pass through the models to generate contextualized feature vectors using NeuroX toolkit \cite{dalvi-etal-2023-neurox}. Subsequently, we apply K-means clustering to the feature vectors, yielding K clusters (also referred to as encoded concepts) for both base and fine-tuned models. We set K = 600 and filter out representations that appear at least 10 times, following the settings prescribed by \cite{hawasly-etal-2024-scaling}. 
We consider an encoded concept to be aligned with the linguistic concept, if it has at least 90\% ($\theta=0.9$) match in the number of words.

\section{Findings and Analysis}
\label{sec:finding}

\subsection{Comparing Modalities}

In Figure \ref{fig:models-alignment}, we illustrate how concepts learned by text, speech, and multimodal models align with the linguistic taxonomies studied in this paper. The alignment patterns across layers reveal distinct processing strategies: speech and text models handle linguistic information differently. Specifically, speech models do not encode word-based linguistic taxonomies in early layers; instead, linguistic structures emerge in the middle layers and peak in the upper layers. This trend is consistent with SpeechT5, which has been jointly trained with text.

Unlike speech models, which gradually transition from acoustic to linguistic representations, text models operate directly on tokenized inputs, allowing them to encode linguistic taxonomies from the initial layers. Text models capture subword-level information such as suffixation early on, while morphology (POS), syntax (chunking), and semantics (SEM) develop progressively, peaking in the middle layers. This pattern suggests that text models incrementally build linguistic understanding, with deeper layers focusing more on integrating syntax and semantics.

Both speech and text models exhibit a falling pattern in alignment in the upper most layers. This decline likely reflects the increasing abstraction of learned representations. In speech models, this suggests a shift from phonetic and word-level patterns to more holistic cues such as paralingusitic 
representations. In text models, the decrease in alignment indicates a transition toward contextual abstraction and task-specific reasoning, where representations become more specialized for downstream tasks rather than directly reflecting linguistic taxonomies.

While SpeechT5-Speech follows the observed trend in speech models, SpeechT5-Text exhibits a different alignment pattern. Unlike BERT and Seamless-text, which refine linguistic representations throughout the network, SpeechT5’s shared encoder allocates less capacity to explicit linguistic taxonomies in its deeper layers. This is due to its multimodal training objective, which aligns speech and text representations for speech-to-text and text-to-speech tasks. Unlike BERT, optimized for masked language modeling and hierarchical linguistic abstraction, and Seamless, where text and speech do not share the latent space, SpeechT5’s encoder is optimized for cross-modal consistency, leading to representations that do not follow typical patterns observed in other models.

Overall, our findings suggest that speech models allocate less capacity to learning linguistic and semantic taxonomies than text models, likely because they must also account for speech-specific features like phonetics, prosody, and speaker variability, which consume much of the network’s representational capacity. As a result, linguistic structures emerge later in speech models and remain less prominent throughout their layers. We speculate that this constraint limits the ability of speech models to encode higher-level conceptual abstractions as effectively as text models, which operate directly on symbolic representations and can dedicate more capacity to linguistic and semantic processing. This distinction could explain why speech models often struggle with tasks requiring deep linguistic reasoning or structured understanding compared to their text-based counterparts \cite{GLUE,SLUE}. 



\subsection{Task-specific Concepts}

\begin{figure}
     \centering
     \begin{subfigure}[b]{0.5\textwidth}
         \centering
         \includegraphics[width=\textwidth]{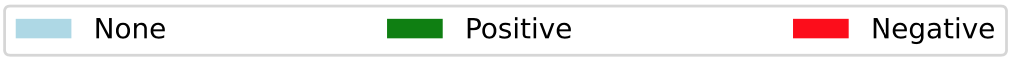}
     \end{subfigure}

      \begin{subfigure}[b]{0.23\textwidth}
         \centering
         \includegraphics[width=\textwidth]{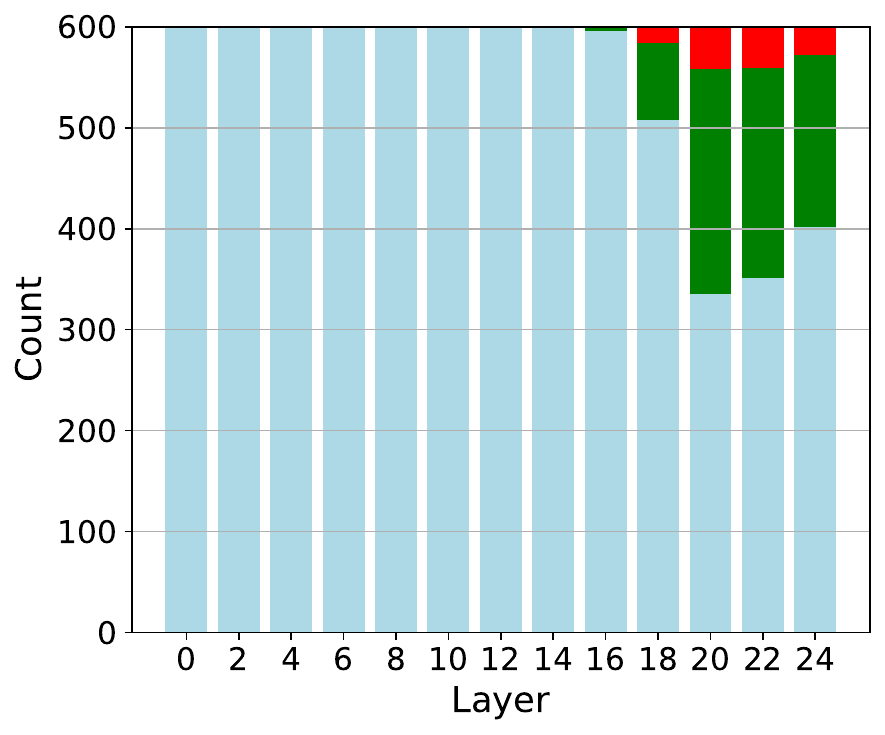}
         \caption{HuBERT--SST }
         \label{fig:hubert-sst-alignment}
     \end{subfigure}
     \begin{subfigure}[b]{0.23\textwidth}
         \centering
         \includegraphics[width=\textwidth]{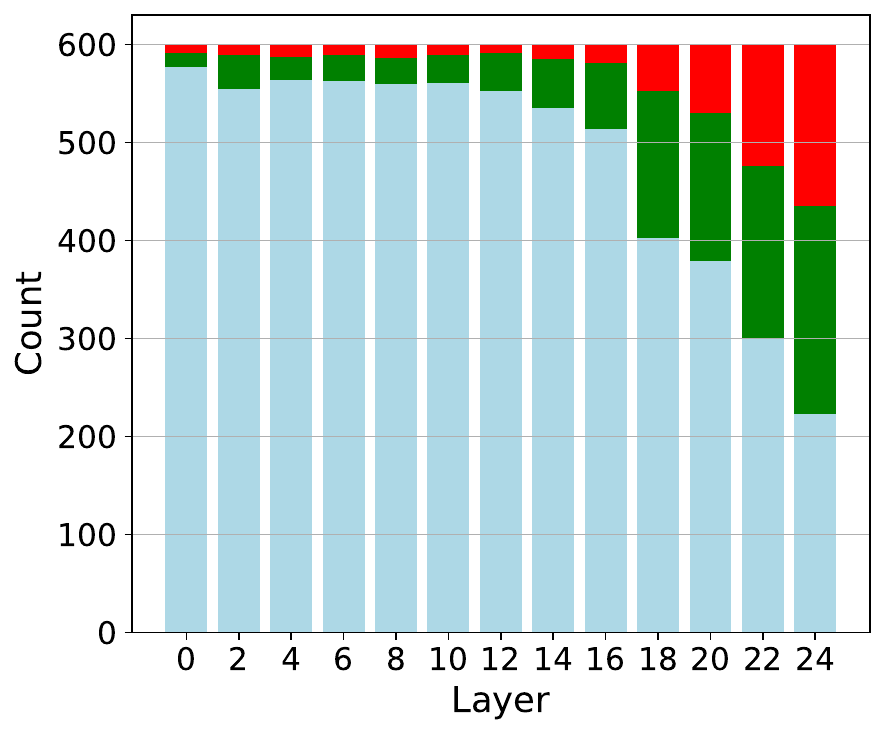}
         \caption{BERT--SST}
         \label{fig:sst-bert-fintuned}
     \end{subfigure}

     \vspace{-0.3cm}
     \caption{Alignment with task-specific polarity concepts
     }
     \vspace{-0.3cm}
    \label{fig:sst-models-alignment}
\end{figure}

In the previous section, we hypothesized that the observed decrease in alignment in the final layers reflects a shift toward task-specific reasoning, where the model's representations become more specialized for downstream tasks. To test this hypothesis, we compared the BERT and HuBERT large models trained for sentiment classification. Using SST2-text and our  SST2-audio, we extracted activation vectors and underlying concepts from the models and aligned them with the output classes: positive and negative. Our results, presented in Figure \ref{fig:sst-models-alignment}, indicate that polarity concepts begin to emerge in the final layers of both the fine-tuned text and speech models.

A closer analysis reveals a notable asymmetry between text and speech-based models in how they encode sentiment. Specifically, the speech-based SST models predominantly rely on signals from positive polarity concepts, whereas textual models exhibit a more balanced distribution of positive and negative concepts. This suggests that speech models may struggle to capture negative sentiment as effectively as textual models, potentially due to the inherent differences in how sentiment is conveyed in speech versus text. In text, explicit negations and sentiment-laden words provide clear cues for polarity, whereas speech-based sentiment relies more on prosodic features such as intonation, pitch, among others which may be harder to disentangle in the absence of large-scale speech-specific supervision.

This observation aligns with task performance: HuBERT underperformed in predicting negative sentiment (accuracy of 87.48\% versus BERT’s 93.21\%), while both models performed comparably when classifying positive sentiment (93.31\% versus 94.98\%). The disparity in negative sentiment prediction highlights a potential limitation of current speech-based sentiment classifiers, which may require additional fine-tuning or explicit modeling of prosodic features to achieve performance parity with text-based models. 
Overall, our results reinforce the idea that final layer representations are shaped by task-specific requirements \cite{merchant-etal-2020-happens,durrani-etal-2021-transfer}, but they also highlight potential gaps in how different modalities encode sentiment. Addressing these gaps could involve integrating additional linguistic or prosodic cues in speech-based models, leveraging multi-modal learning strategies, or exploring alternative architectures that better capture the nuances of spoken sentiment.


\subsection{Qualitative Analysis}

In our qualitative analysis, we explored how the models learn and represent concepts by leveraging GPT for annotation. While we identified concepts that align with established linguistic taxonomies, it's important to note that these models may not always strictly conform to human-defined concepts. To interpret and analyze these concepts more effectively, we can annotate them for further exploration. Following \cite{mousi2023can}, we employ ChatGPT in a zero-shot setting, prompting the model with structured instructions, reported in Listing \ref{lst:prompt_for_answer}. 


\noindent Our findings suggest that concepts in the models are often organized in compositional hierarchies, where the model initially groups concepts based on a primary objective and then refines them according to semantic relations. For example, the model may first group all names starting with the sound \textipa{dZ} before distinguishing between other linguistic or semantic categories (see the concept in Figure \ref{fig:HubertL12c141}). This hierarchical structure highlights the model's ability to form complex, layered associations among concepts, which may offer insights into how it processes and organizes knowledge.

\begin{lstlisting}[style=mystyle,caption={Prompt for label assignment.},label={lst:prompt_for_answer}]
Assistant is a large language model trained by OpenAI.
Instructions:
Give a short and concise label that best describes the following list of words:
["w_1", "w_2", ..., "w_N"]
\end{lstlisting}
\vspace{-0.2cm}

\section{Related Work} 
The discovery and interpretation of latent concepts in deep models remain crucial challenges in NLP, particularly in speech processing. Recent studies have focused on understanding the internal representations learned by these models, with an emphasis on layer-wise analysis and latent concept discovery~\cite{tenney-etal-2019-bert,jawahar-etal-2019-bert,michael-etal-2020-asking,sajjad-etal-2022-analyzing,durrani-etal-2022-latent}. These foundational works have explored how hierarchical linguistic features--such as surface properties, syntax, and semantics are distributed across layers. The findings suggest a progression from lexical and syntactic features in the lower and middle layers to more abstract semantic representations in the higher layers. 

There are a handful of studies on how information is encoded in speech models \cite{chowdhury2024end,waheed2024speech, el2024speech,lee2024multimodal,de2024human}; however, latent concept discovery and the evolution of representations across layers remain largely underexplored compared to text-based models.
Studies such as \cite{martin23_interspeech,cho2024sd,wells22_interspeech} indicate that phonetic and phonemic distinctions emerge in the early layers, often overshadowing semantic information in speech models \cite{choi2024self}. Previous analyses show a strong alignment between HuBERT's latent units and linguistic structures, particularly phonetic categories \cite{wells22_interspeech}, and syllabic \cite{cho2024sd}. Additionally, studies like \cite{pasad2024self} demonstrate that model size and training objectives significantly influence the distribution of linguistic information.  Despite these insights, in-depth research on speech, multimodal, and encoder-decoder models remains underexplored. To address these gaps, our study focuses on understanding how semantic concepts emerge in deep speech models.

\section{Conclusion}

In this study, we compared speech, text, and multimodal models to understand how they represent linguistic concepts. Our findings suggest that text models, such as BERT, directly encode linguistic structures from early layers, while speech models, like HuBERT, gradually develop linguistic representations from acoustic features. Multimodal models like SpeechT5 exhibit unique alignment patterns due to cross-modal training. We observed that speech models allocate less capacity to linguistic taxonomies, focusing more on speech-specific features like phonetics. In task-specific tests, such as sentiment analysis, speech models showed weaker performance in capturing negative sentiment compared to text models. Our results emphasize the different ways these models process and internalize language, with text models offering richer, more structured linguistic representations. 

\bibliographystyle{IEEEtran}
\bibliography{mybib}

\end{document}